\documentclass[conference]{IEEEtran}
\usepackage{cite}
\usepackage{amsmath,amssymb,amsfonts}
\usepackage{algorithmic}
\usepackage{graphicx}
\usepackage{textcomp}
\usepackage{xcolor}
\usepackage{array}
\usepackage{fixltx2e}
\makeatletter
\let\MYcaption\@makecaption
\makeatother
\usepackage[font=footnotesize]{subcaption}
\makeatletter
\let\@makecaption\MYcaption
\makeatother

\def\BibTeX{{\rm B\kern-.05em{\sc i\kern-.025em b}\kern-.08em
    T\kern-.1667em\lower.7ex\hbox{E}\kern-.125emX}}
\begin{document}

\title{Stream-Flow Forecasting of Small Rivers Based on LSTM\\
}

\author{\IEEEauthorblockN{Youchuan Hu}
\IEEEauthorblockA{\textit{College of Letters and Science} \\
\textit{University of California, Los Angeles}\\
Los Angeles, United States \\
aaronndx@ucla.edu}
\and
\IEEEauthorblockN{2\textsuperscript{nd} Le Yan}
\IEEEauthorblockA{\textit{Computer and Information College} \\
\textit{Hohai University}\\
Nanjing, China \\
yanle@hhu.edu.cn}
\and
\IEEEauthorblockN{3\textsuperscript{rd} Tingting Hang}
\IEEEauthorblockA{
\textit{Wanjiang University of Technology}\\
Maanshan, China \\
httsf@hhu.edu.cn}
\and
\IEEEauthorblockN{4\textsuperscript{th} Jun Feng}
\IEEEauthorblockA{\textit{Computer and Information College} \\
\textit{Hohai University}\\
Nanjing, China \\
fengjun@hhu.edu.cn}
}

\maketitle

\begin{abstract}
Stream-flow forecasting for small rivers has always been of great importance, yet comparatively challenging due to the special features of rivers with smaller volume. Artificial Intelligence (AI) methods have been employed in this area for long, but improvement of forecast quality is still on the way. In this paper, we tried to provide a new method to do the forecast using the Long-Short Term Memory (LSTM) deep learning model, which aims in the field of time-series data. Utilizing LSTM, we collected the stream flow data from one hydrologic station in Tunxi, China, and precipitation data from 11 rainfall stations around to forecast the stream flow data from that hydrologic station 6 hours in the future. We evaluated the prediction results using three criteria: root mean square error (RMSE), mean absolute error (MAE), and coefficient of determination (R\textsuperscript{2}). By comparing LSTM's prediction with predictions of Support Vector Regression (SVR) and Multilayer Perceptions (MLP) models, we showed that LSTM has better performance, achieving RMSE of 82.007, MAE of 27.752, and R\textsuperscript{2} of 0.970. We also did extended experiments on LSTM model, discussing influence factors of its performance.
\end{abstract}


\section{Introduction}

Worldwide, floods are considered as one of the most common and naturally distributed risks to life and property \cite{balica2013parametric}. According to \cite{wallemacq2017disaster}, in 2017, flood were the most influential disaster with respect to number of people affected - 59.6\% of people affected by natural disasters were affected by flood. Due to its burstiness and uncertainties, floods remain to be a comparatively hard-to-prevent disaster, and more advanced controls methods are eagerly needed. Among them, flood forecasting is always a crucial one. A timely and precise advance warning allows ample time for more mitigating actions and less damage by the disaster. However, when it comes to medium or small rivers, various problems exert excess challenges on the forecasting process. Due to the low capacity of those rivers, floods often abrupt in appearance, rapid in confluence, and short in forecast period \cite{feng2018hydrologic}. Thus, more sophisticated forecast methods are always in high demand.

Main traditional stream flow forecasting methods are those which employ physical hydrologic models or traditional machine learning algorithms. Hydrologic models that use the data of river stage, stream flow, or runoff volumes to forecast floods are mainly based on mathematical and physical analysis of hydrologic process, thus they are usually deterministic, and forecast results are normally exhibited as time series of estimates \cite{han2017bayesian}. Results of these models are often easily deteriorated if the data fed in contain certain degree of error or environmental noise \cite{damavandi2019accurate}. With the development of artificial intelligence and the approach of big data, researchers began to use ‘data-driven’ models – instead of mathematic or physic models - to study various aspects of hydrological phenomenon. ‘Data-driven’ models focus less on the exact logic and physic theories behind the forecast and more on the potential relationships lying inside the huge amount of data, thus remarkably reduce the amount of work done due to the non-linear feature and noise complexity of hydrological models, and improve the accuracy of the forecast. However, traditional machine learning models manipulate every input and output in a discrete manner, thus have limited performance in the area of prediction, which involves time-series data and every piece of data at a certain time has relationship with recent data. Prediction result of most of traditional models are accompanied by considerable errors.

As described above, physical models and traditional machine learning algorithms both have limited performance due to: 1, erroneous and chaotic data, and 2, special feature of time-series prediction. In this paper, we use the LSTM (Long Short-Term Memory) model – a kind of circular memory neural networks developed from RNN (Recurrent Neural Network) – in stream flow prediction to try to solve above two problems. As a sophisticated machine learning model, LSTM works well in dealing with chaotic data resulting from the complexity of real environment, and instability of medium or small rivers. Moreover, our prediction method involving the use of LSTM has innovation comparing to methods using traditional models, in the way that LSTM has “memory”, and every output is based on previous outputs, thus has ability to take advantage of the information “between” time-series data, and works better in predicting the stream flow changes that is a trend along time. The main idea of this paper is to use LSTM to analysis big amount of stream flow data, accompanied by rainfall data collected from various precipitation stations along the rivers to estimate the future stream flow of a certain spot in a river, and compare the prediction result with traditional machine learning model SVR (Support Vector Regression) and deep learning model MLP (Multilayer Perceptions). The results of the comparative experiment conducted in this paper proved that LSTM model contributes to stream-flow forecasting of small rivers with respect to:

\begin{enumerate}
    \item Better model stability. Different from other two models, LSTM performs forecast that does not produce frequent and obvious fluctuations of stream flow line in cases of small rainfalls.
    \item Better model reliability. LSTM is more accurate in forecasting stream flow peaks, which is vital to early warning of floods.
    \item More intelligent in capturing the features of data. By extended experiments, we observed that LSTM is able to read different combinations of input data, including history stream flow volume, rainfall data, and areal rainfall data, and improve model accuracy based on all of them.
\end{enumerate}

The rest of this paper is organized as follows. In section 2, works related to development and current situation of stream flow forecasting are listed. In section 3, the RNN model, which is the origin and foundation of LSTM, and the LSTM model are introduced. Then the complete experiment process of testing the performance of LSTM, including data preparation, model training, comparative models selecting, evaluation criteria choosing, final results, and extended experiments of LSTM performance are presented in section 4. At last, conclusion comes out in section 5.

\section{Related Works}

In recent years, there are more and more data-driven AI model stream flow forecasting methods that are developed and put into practice. According to Yaseen, \textit{et al.} \cite{yaseen2015artificial}, internationally there are mainly five areas of focus: ANN (Artificial neural network), SVM (Support vector machine), Fuzzy (Fuzzy logic method), EC (Evolutionary computing methods), and W-AI (Wavelet-complementary modeling).

An ANN is a kind of Artificial intelligence information processing system that resembles the biological neural networks of human brains \cite{haykin1994neural}. In 2002, Hsu \textit{et al.}\cite{hsu2002self} proposed the self-organizing linear output map (SOLO) – a kind of multivariate ANN procedure – to forecast rainfall-runoff. Cigizoglu \cite{cigizoglu2005application} tested the performance of GRNN (Generalized regression neural network) regarding the intermittent daily mean flows forecasting and estimation in 2005.  In 2010, Kagoda \textit{et al.} \cite{kagoda2010application} used RBFNN (Radial Basis Function Neural Network) to perform 1-day forecasts of stream-flow and proved that it is a relatively more superior method. 

SVM is popularized in last 20 years as an effective method solving the noisy problems. In 2005, Sivapragasam and Liong \cite{sivapragasam2005flow} experimented the performance of SVM in stream-flow prediction and yielded promising results.  Asefa \textit{et al.} \cite{asefa2006multi} used SVM approach to predict seasonal and hourly multi-scale stream-flow in 2006.  In 2011, Noori \textit{et al.} \cite{noori2011assessment} assessed the input variables determination on the SVM model performance using PCA, Gamma test, and forward selection techniques for monthly stream flow prediction. 

The theory of fuzzy sets was introduced by Lotfi A. Zadeh in 1965. Fuzzy has been used to deal with the uncertainties inside the variables in models. In 2007, a neuro-fuzzy model was introduced by El-Shafie \textit{et al.} \cite{el2007neuro} to forecast the monthly basis inflow of the Nile river.  In 2009, \"{O}zger \cite{ozger2009comparison} utilized the Mamdani and the Takagi–Sugeno (TS) fuzzy inference systems for stream-flow value prediction.  Sanikhani and Kisi \cite{sanikhani2012river}, in 2012, developed two different adaptive neuro-fuzzy (ANFIS) techniques to estimate monthly river flow. 

EC (Evolutionary Computing) is the collective of Evolutionary Algorithms (EA) that are used in the process of selection, mutation, and reproduction on a population of individual structures that undergo evolution \cite{han2017bayesian}.  In 1999, Savic \textit{et al.} \cite{savic1999genetic} conducted the first research on the employment of Evolutionary Computing in the field of stream-flow modeling.  The performance of Genetic Programming and ANN in stream-flow forecasting were compared by Makkeasorn \textit{et al.} \cite{makkeasorn2008short} in 2008.  In 2009, the river inflow prediction ability of LGP was investigated by Guven \cite{guven2009linear} and the comparison with MLP and GRNN methods was carried out, and the result proved that LGP had a better performance. 

Wavelet Transform (WT) is a method that focuses on handling data of time series. Wavelet and neuro-fuzzy conjunction model was employed by Shiri and Kisi \cite{shiri2010short} in 2010 to make daily, monthly, and yearly stream-flow model.  In 2014, wavelet transform-genetic algorithm-neural network model (WAGANN) was proposed by Sahay and Srivastava \cite{sahay2014predicting} for forecasting monsoon river flows one day ahead.

\section{Models}

\subsection{Recurrent Neural Network (RNN)}\label{AA}
First developed in 1980s, RNN obtained its specialness due to its structure: the neurons are connected with each other and self-looped, thus the structure is able to display dynamic temporal behaviors and “remember” the information from last process \cite{zhang2018developing}. 
The basic and classic logic of RNN is presented below \cite{pascanu2013construct}: 
\begin{equation}
h_t = f_h(x_t,h_{t-1}) = \phi_h(W^Th_{t-1} + U^Tx_t)\label{eq}
\end{equation}
\begin{equation}
y_t = f_o(h_t,x_t) = \phi_o(V^Th_t)\label{eq}
\end{equation}

In one unit, \(x_t\) is the input, ht represents the hidden state, and \(y_t\) is the output. The subscript represents time. Firstly, hidden state output from last time is combined with current input (each with the weights \(W^T\) and \(U^T\)), the result of which is transformed by a nonlinear function - \(\tanh\) or sigmoid, conventionally – and then fed into the hidden state. Then, the hidden state takes its weight \(V^T\), transformed by another nonlinear function, and at last the result is accepted by \(y_t\). In this way, current output \(y_t\) is affected by last hidden state, thus obtains “short memory”. 

One significant problem of classical RNN is that, due to its looped feature, the error of backward propagation depends on the weights in an exponential manner. Thus, error signals of RNN vanish or blow up in long-term process \cite{schmidhuber1997long}.

\subsection{Long Short-Term Memory network (LSTM)}
To solve the gradient blowing up or vanishing problem, LSTM was introduced by Hochreiter and Schmidhuber \cite{schmidhuber1997long} in 1997, which used memory cells and gates to control the long-term information saved in the network – keep or through away.
\begin{equation}
g_t = \sigma(U_gx_t + W_gh_{t-1} + b_f)\label{eq}
\end{equation}
\begin{equation}
i_t = \sigma(U_ix_t + W_ih_{t-1} + b_i)\label{eq}
\end{equation}
\begin{equation}
\widetilde{c_t} = \tanh(U_cx_t + W_ch_{t-1} + b_c)\label{eq}
\end{equation}
\begin{equation}
c_t = g_t * c_{t-1} + i_t * \widetilde{c_t}\label{eq}
\end{equation}
\begin{equation}
o_t = \sigma(U_ox_t + W_oh_{t-1} + b_o)\label{eq}
\end{equation}
\begin{equation}
h_t = o_t * \tanh(c_t)\label{eq}
\end{equation}

$U$ and $W$ are the weights of input into different gates: input gate (\(i_t\)), input modulate gate (\(\widetilde c_t\)), forget gate (\(g_t\)), and output gate (\(o_t\)). $b$ is bias vectors, \(c_t\) is cell state, and \(h_t\) is hidden state. All these controllers determine how much information to receive from the last loop, and how much to pass to the new state.

By actively choosing useful information to store and others to reject, LSTM provides a solution to the gradient explosion and vanishing problem faced by RNN.

\section{Experiments}

In this section, the complete experiment process of the stream-flow forecast of the rivers in Tunxi, China using Artificial Intelligence data-driven model is presented, including data preparation, model training, comparative models selecting, evaluation criteria choosing, final results, and expended experiments of LSTM performance.

\subsection{Data Collection and Division}

The data for the experiment is collected from Tunxi District, Huangshan City, Anhui Province, China. According to \cite{dong2015flood}, Tunxi catchment has a drainage area of 2696.76 km\textsuperscript{2}. Its altitude is low in east and increases gradually towards west. As affected by continental monsoon climate, the rainfall differs a lot between years. In one year, the rainfall also has uneven separation. More than 50\% of the annual precipitation happens between April and June. Stream-flow changes in Tunxi area have the feature of small rivers: complexity and abruptness, which is suitable to test the forecast ability of models.

The experiment data consists of the stream-flow volume data of Tunxi which was collected from a hydrologic station, and rainfall data from 11 precipitation stations located on the upstream of the hydrologic station.
There are in total 18648 pieces of data collected from 1981 to 2003.

\subsection{Data Pre-processing}

The experiment will use the stream-flow data and precipitation data from 11 rainfall stations in the past 12 hours to forecast the stream-flow volume of the 6\textsuperscript{th} hour in the future. In order to transform the raw data into the form suitable for supervised learning, in this experiment, a series\_to\_supervised function is used. After the transformation, the data turns into the form as shown in Tab.~\ref{datatrans}.

Q(t+X) represents the stream-flow data from (t-12) to (t+5), which means from 12 hours in the past to 6 hours in the future. P\textsubscript{1}(t+X) to P\textsubscript{11}(t+X) represents the precipitation data of the 11 rainfall stations from (t-12) to (t+5). Then, the 1\textsuperscript{st}-144\textsuperscript{th} columns (from Q(t-12) to P\textsubscript{11}(t-1)) are selected to be the features (x set), which contain the stream-flow and all the precipitation data of the past 12 hours. The 205\textsuperscript{th} column (Q(t+5)) is selected to be the target (y set), which is the stream-flow data of the 6\textsuperscript{th} hour in the future.

When the data are transformed into time-series format, only those which have enough data in front of and after it to form a time series are kept. Thus, some rows at the beginning or in the end are thrown away. After the 12 - 6 transformation mentioned above, 18237 pieces of data are kept. They are divided by an around 7:3 ratio - 13000 pieces are used as a training set, and 5237 used as a test set.

As the data of this experiments is collected from different stations and through a large time span, the dimension of the different sets of data are not the same. In order for the models to have better performance, the data goes through a normalization process using the MinMaxScaler function in the sklearn package, and is unified to [0,1]. The formula follows:

\begin{equation}
x^{'} = \frac{x-\min(x)}{\max(x) - \min(x)}\label{eq}
\end{equation}

\subsection{Model Training}

\begin{figure}[t]
\centering
\includegraphics[width=0.4\textwidth]{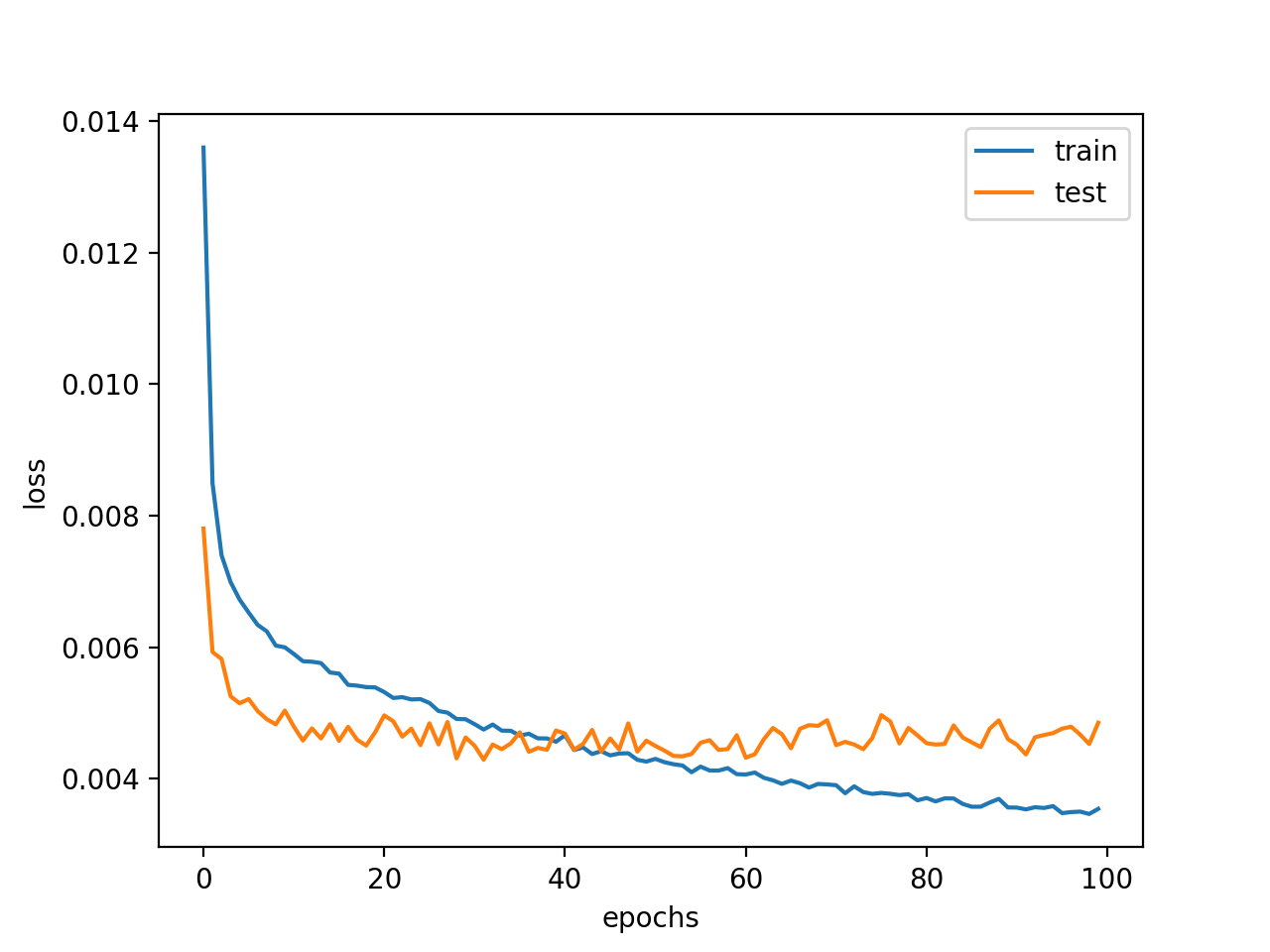}
\caption{Loss of train and test sets of LSTM models with different epochs.}
\label{epochs}
\end{figure}

The LSTM model used in the experiment is based on keras library, the python deep learning library. 
The amount of hidden layer nodes is one of the parameters need to be determined in the model. By experiments, model with 64 nodes has the best performance.
The optimizer, batch size and epochs are also parameters that influence the performance of the model. The choice of optimizer influences how the loss function is minimized, thus how the model heads to the final outcome. Standard choices include momentum, Adagrad, RMSProp, Adam, etc. By experiments, the Adam optimizer is chosen. Batch size affects the amount of data processed at a time. Through batches, the model updates multiple times before processing the whole dataset and thus the dynamics of the process is affected. As small batch size greatly slows down training speed and big batch size causes overfitting, on balance the batch size is set to 72 in this experiment. Epochs are the times the model runs through the whole data. According to Fig.~\ref{epochs}, when epochs are approximately 30, the loss of test set is the lowest. So, the epochs are set to 30 in this experiment.

\begin{table}[t]
\caption{Headings of Data After Series\_to\_Supervised Transformation}
\begin{center}
\begin{tabular}{ |c|c|c|c|c|c|c|c| } 
 \hline
 \multicolumn{5}{|c|}{Features} & & Target & \\
 \hline
 Q(t-12) & P\textsubscript{1}(t-12) & P\textsubscript{2}(t-12) & ... & P\textsubscript{11}(t-1) & ... & Q(t+5) & ... \\ 
 \hline
\end{tabular}
\end{center}
\label{datatrans}
\end{table}

\subsection{Comparative Models Selecting}
To evaluate the performance of the proposed model LSTM, another two models are chosen to be comparative models. The former is a traditional machine learning method, while the latter is a deep learning method.
\begin{itemize}
\item SVR: Support Vector Regression is a model derived from Support Vector Machine. According to \cite{basak2007support}, "The idea of SVR is based on the computation of a linear regression function in a high dimensional feature space where the input data are mapped via a nonlinear function." The kernel function and two parameters - C and gamma - should be determined for model setup. In this paper, RBF kernel function is selected, and by grid search, the combination of (C=0.095, gamma=0.165) is chosen.
\item MLP: Multilayer perceptrons are a class of ANN, which the nonlinear computing elements are arranged in a feed-forward layered structure \cite{bourlard1988auto}. In this paper, the MLP model of 1 hidden layer is selected.
\end{itemize}

\subsection{Evaluation Criteria}

Three metrics are used in this paper as the evaluation criteria: root mean square error (RMSE), median absolute error (MAE), and coefficient of determination (R\textsuperscript{2}).

RMSE is a common measurement method to show the difference between value predicted and value observed. Its formula is the following:
\begin{equation}
RMSE = \sqrt{\frac{1}{m}\sum_{i=1}^m(y_i - \hat{y_i})^2}\label{eq}
\end{equation}
Where $m$ denotes the total number of values, \(y_i\) denotes the value predicted, and \(\hat{y_i}\) denotes the value observed. The square root uniforms the outcome (error) scale with the input scale. RMSE value is always non-negative. A lower RMSE value means a better prediction.

MAE works in a similar way to RMSE except that the error is linear. Its formula is the following:
\begin{equation}
MAE = \frac{1}{m}\sum_{i=1}^m\lvert y_i - \hat{y_i} \rvert\label{eq}
\end{equation}
Since it works in a linear way, MAE does not penalize big errors more than small errors, but present them as they were. Similar to RMSE, MAE value is always non-negative, and a lower MAE value means a better prediction.

R\textsuperscript{2}, or coefficient of determination, is a metric based on MSE (MSE is the square of RMSE). It differs from the preceding two metrics in that the scale of outcome does not depend on scale of input. The formula is the following:
\begin{equation}
R^2 = 1 - \frac{\sum_{i=1}^{m}(y_i - \hat{y_i})^2}
{\sum_{i=1}^{m}(y_i - \overline{y})^2}\label{eq}
\end{equation}
\(\overline{y}\) is the mean of all values predicted. The denominator is the total variation of the predicted values. In most of the cases, R\textsuperscript{2} value is in range [0,1], and a higher value means a better prediction.

\subsection{Results}

Feed the data and run the models, the results in the form of errors of SVR, MLP, and LSTM model are in Tab.~\ref{TabResult}.

\begin{figure*}[t]
\begin{center}
\begin{subfigure}{0.32\textwidth}
\begin{center}
\includegraphics[width=\linewidth, height=5cm]{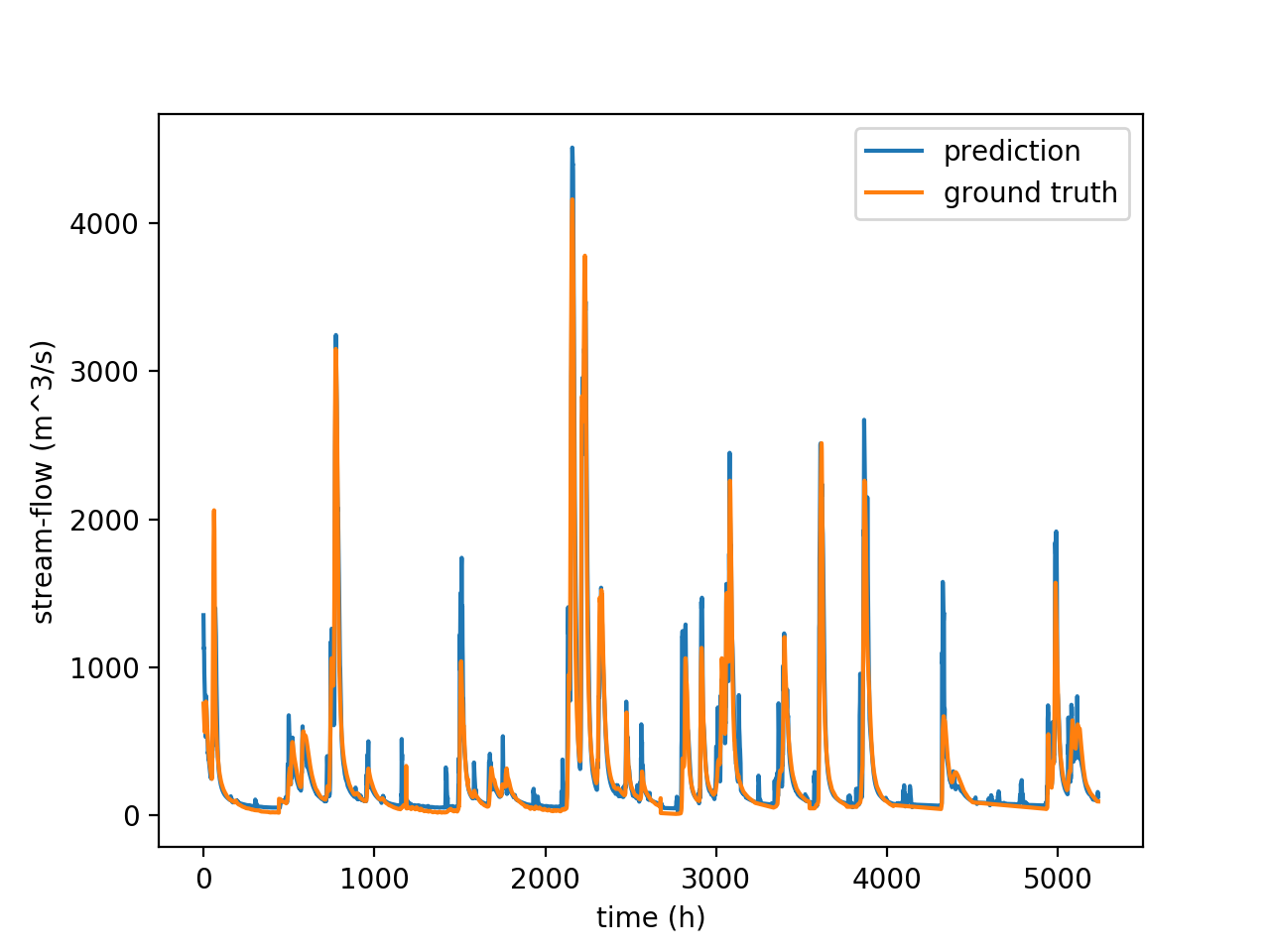} 
\caption{SVR}
\end{center}
\label{svrResult}
\end{subfigure}
\begin{subfigure}{0.32\textwidth}
\begin{center}
\includegraphics[width=\linewidth, height=5cm]{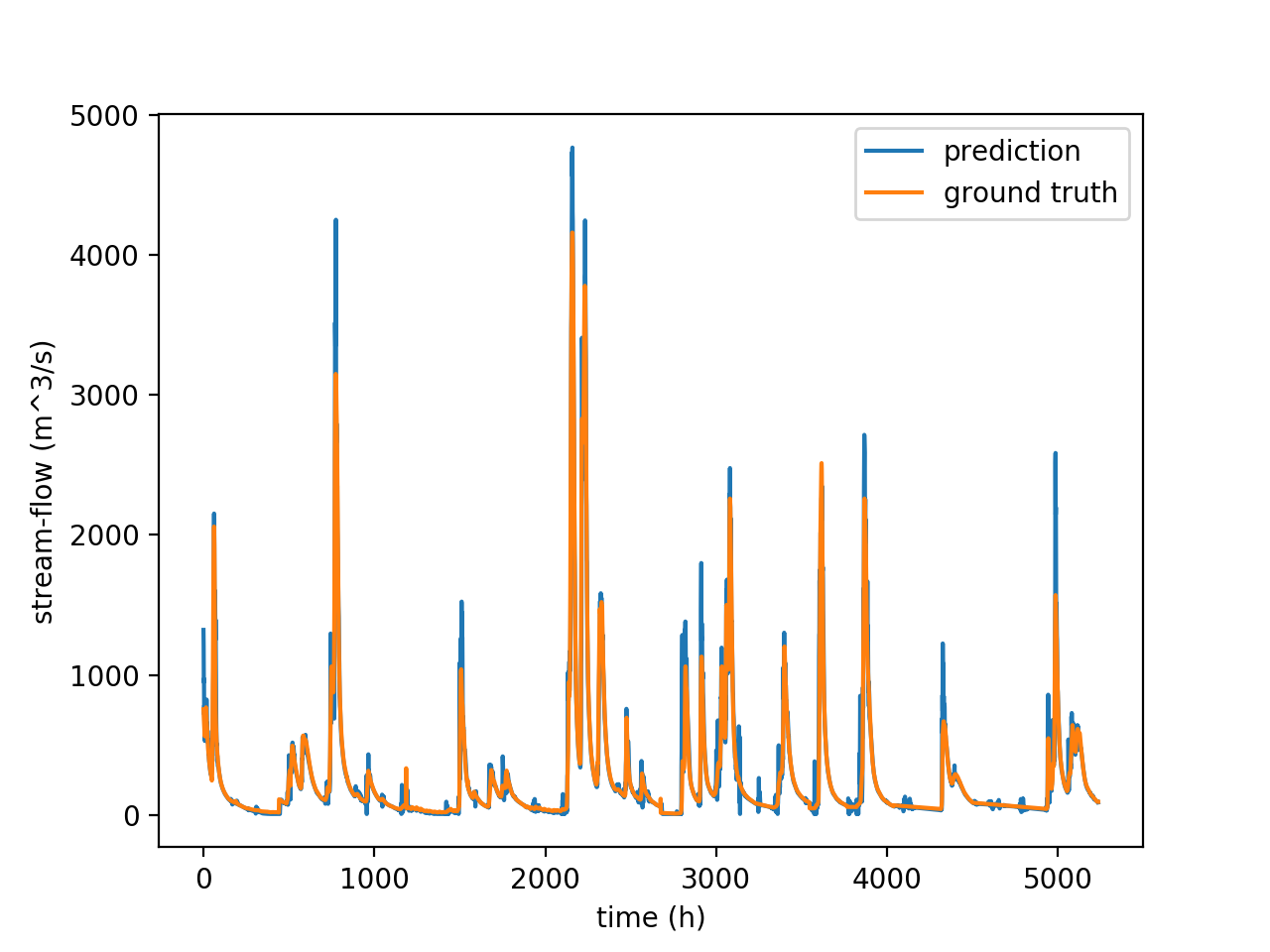}
\caption{MLP}
\end{center}
\label{mlpResult}
\end{subfigure}
\begin{subfigure}{0.32\textwidth}
\begin{center}
\includegraphics[width=\linewidth, height=5cm]{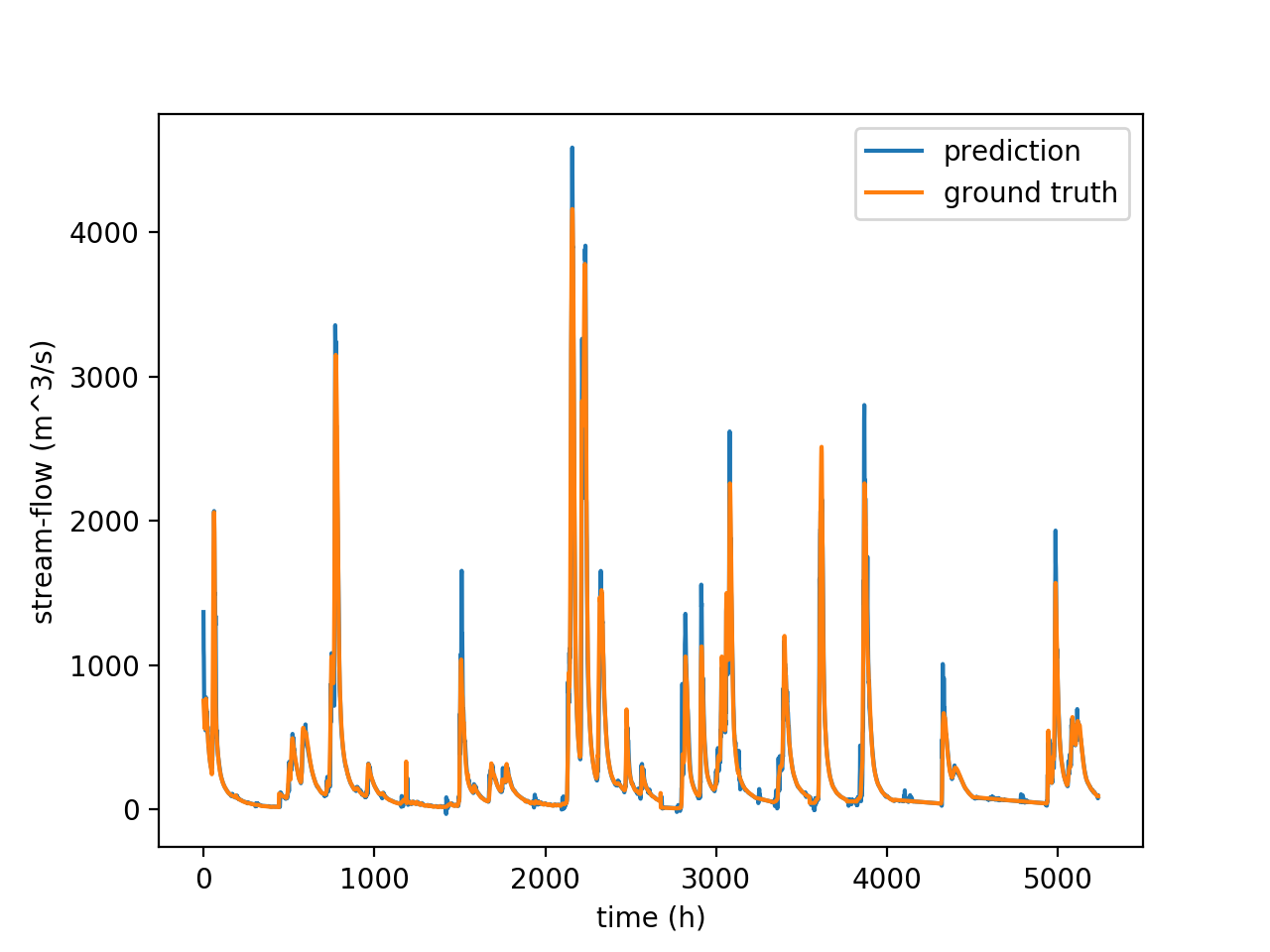}
\caption{LSTM}
\end{center}
\label{lstmResult}
\end{subfigure}
\caption{Prediction results of SVR, MLP, and LSTM models}
\label{FigResult}
\end{center}
\end{figure*}

The prediction results in the form of graphs are in Fig.~\ref{FigResult}.

From Tab.~\ref{TabResult}, we can figure out that LSTM has the best performance among the three models with respect to all three evaluation criteria. Thus, statistically the LSTM model has the best prediction accuracy. Fig.~\ref{FigResult} shows that the LSTM prediction result of stream-flow data almost excellently fits the actual situations. Different from the SVR and MLP predictions, LSTM prediction does not yield obvious nonexistent small peaks or valleys. Moreover, with respect to prediction of major stream flow peaks, LSTM model is considerably better than MLP model, and slightly better than SVR model.

The results show that the remember-forget ability of LSTM greatly helps the model to predict non-linear and time-series data and have a relatively better performance on the forecast of stream-flow of rivers. However, LSTM still have errors in the major peak prediction – most of the major peak predictions exceeds the actual value by approximately 10 per cent. Better results may be achieved through adjustment of training process or larger and better available data base.

\subsection{Extended Experiments for LSTM}

\subsubsection{Combinations of input data}
Tab.~\ref{tab:inputcombs} shows the error of LSTM models fed with different combinations of input data, while all other conditions stay the same as in the standard experiment. The result shows that history stream-flow data play a significant role in the accuracy of forecast, but rainfall (and areal rainfall) data are also indispensable. However, upon the presence of rainfall-related data, different combinations of rainfall data types do not pose a large difference on the result. Rainfall data are relatively more helpful than areal rainfall data.

\subsubsection{Change of predict time step}
\begin{figure*}[t]
\begin{center}
\begin{subfigure}{0.32\textwidth}
\begin{center}
\includegraphics[width=\linewidth, height=5cm]{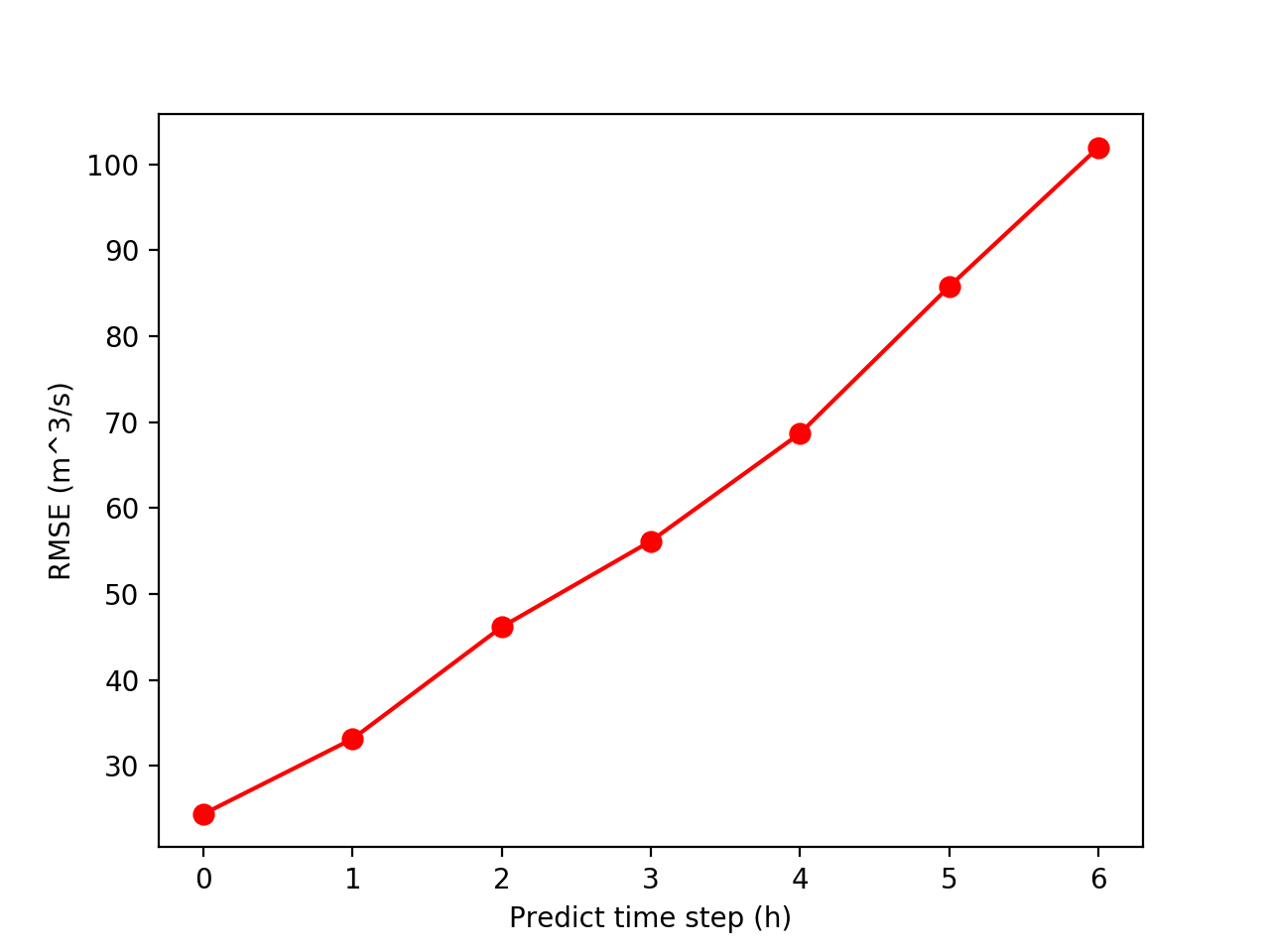} 
\caption{RMSE}
\end{center}
\label{ptsrmse}
\end{subfigure}
\begin{subfigure}{0.32\textwidth}
\begin{center}
\includegraphics[width=\linewidth, height=5cm]{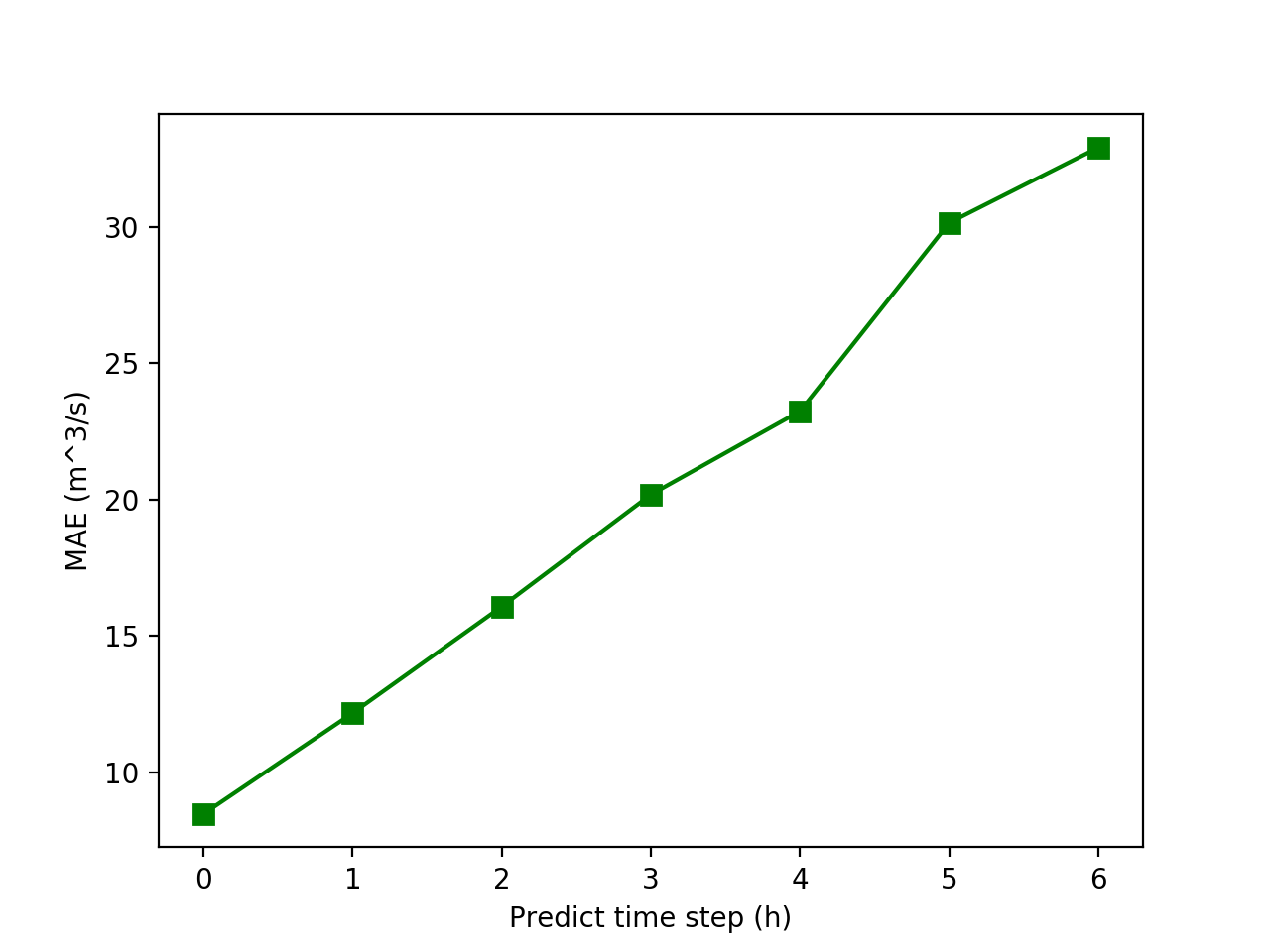}
\caption{MAE}
\end{center}
\label{ptsmae}
\end{subfigure}
\begin{subfigure}{0.32\textwidth}
\begin{center}
\includegraphics[width=\linewidth, height=5cm]{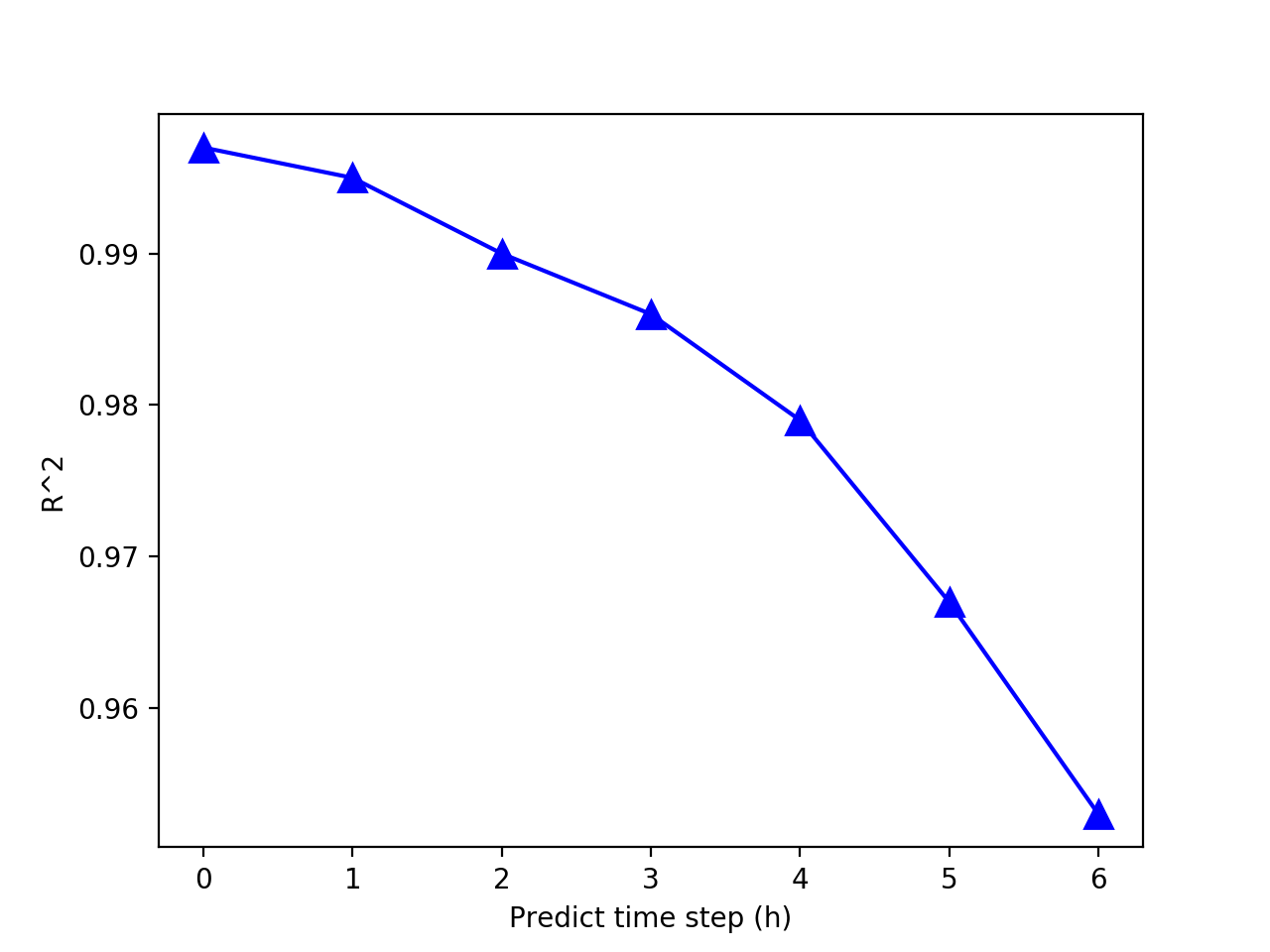}
\caption{R\textsuperscript{2}}
\end{center}
\label{ptsr2}
\end{subfigure}
\caption{Errors of LSTM models with different predict time steps}
\label{PTS}
\end{center}
\end{figure*}

The predict time step is how far in the future does the LSTM model predict. The standard model in this paper has a predict time step of 6. That is, upon receiving the newest data, the model gives out predictions for the 6\textsuperscript{th} hour in the future from now.
Fig.~\ref{PTS} shows the values of three evaluation criteria of LSTM models with different predict time step, while all other conditions stay the same as the standard model in this paper. The results imply that predict time step has a negative correlation with the accuracy of the model, which makes sense since it's harder for models to predict further into the future.

\subsubsection{Change of encoder time step}
\begin{figure*}[t]
\begin{center}
\begin{subfigure}{0.32\textwidth}
\begin{center}
\includegraphics[width=\linewidth, height=5cm]{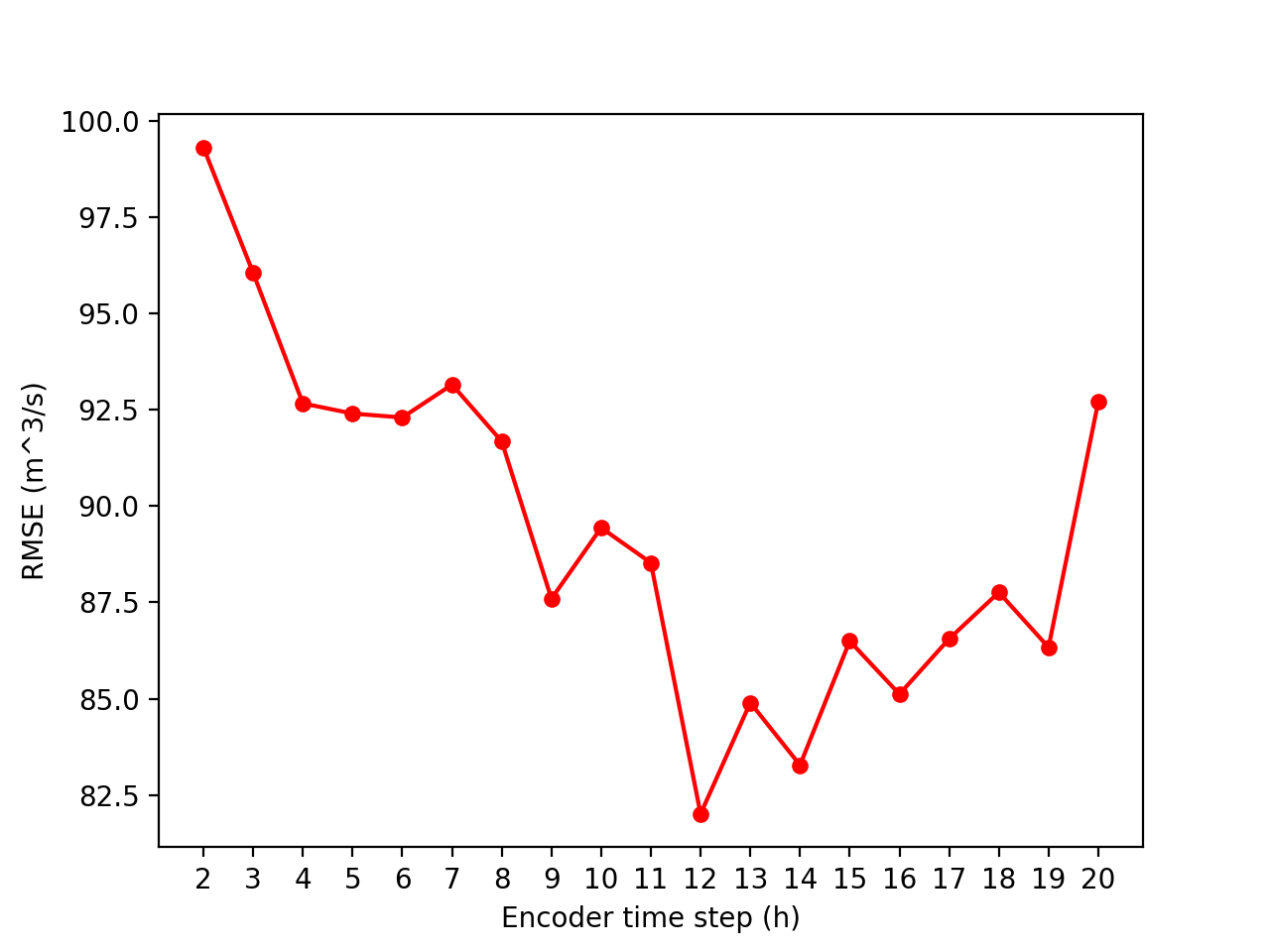} 
\caption{RMSE}
\end{center}
\label{etsrmse}
\end{subfigure}
\begin{subfigure}{0.32\textwidth}
\begin{center}
\includegraphics[width=\linewidth, height=5cm]{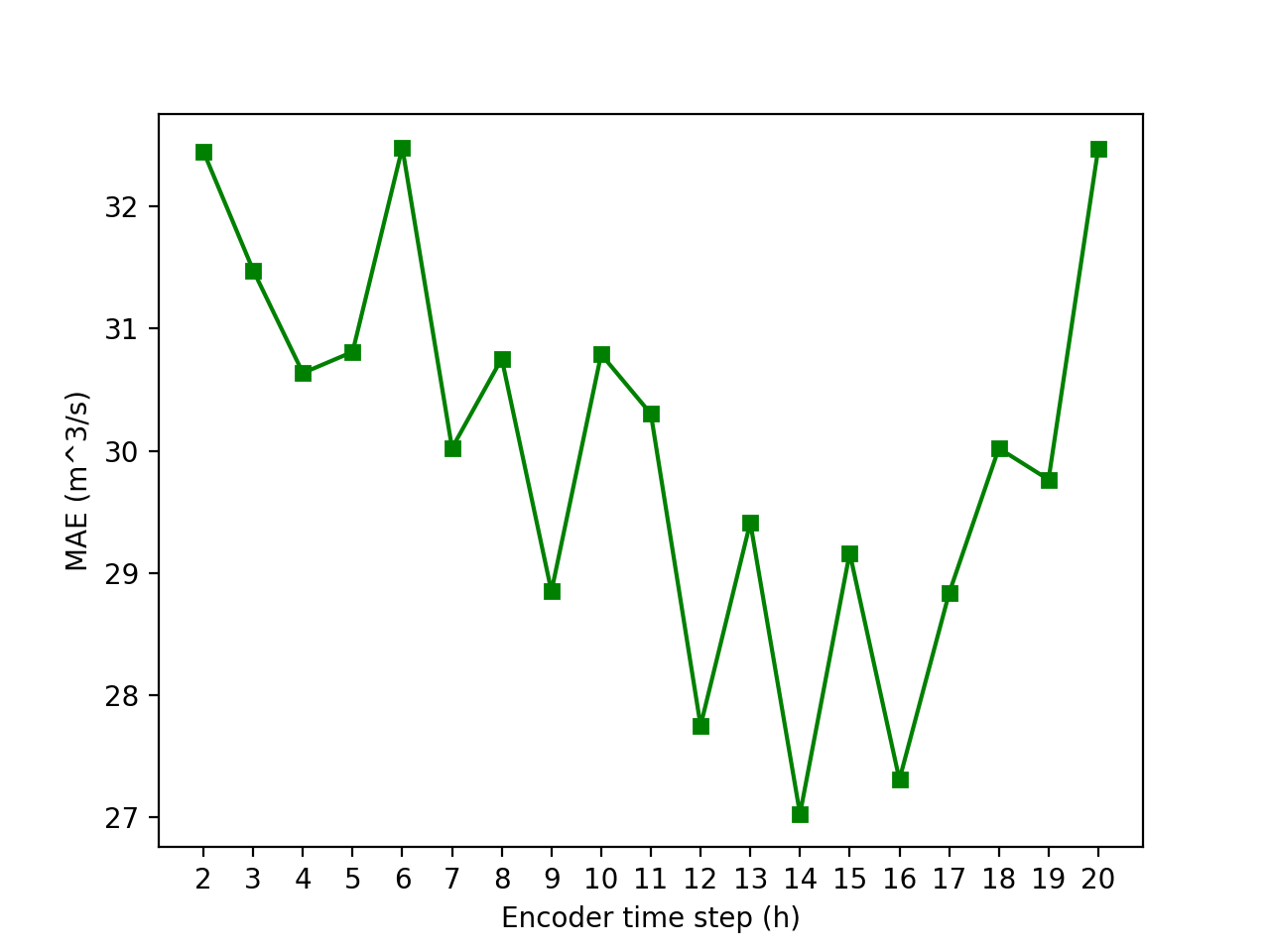}
\caption{MAE}
\end{center}
\label{etsmae}
\end{subfigure}
\begin{subfigure}{0.32\textwidth}
\begin{center}
\includegraphics[width=\linewidth, height=5cm]{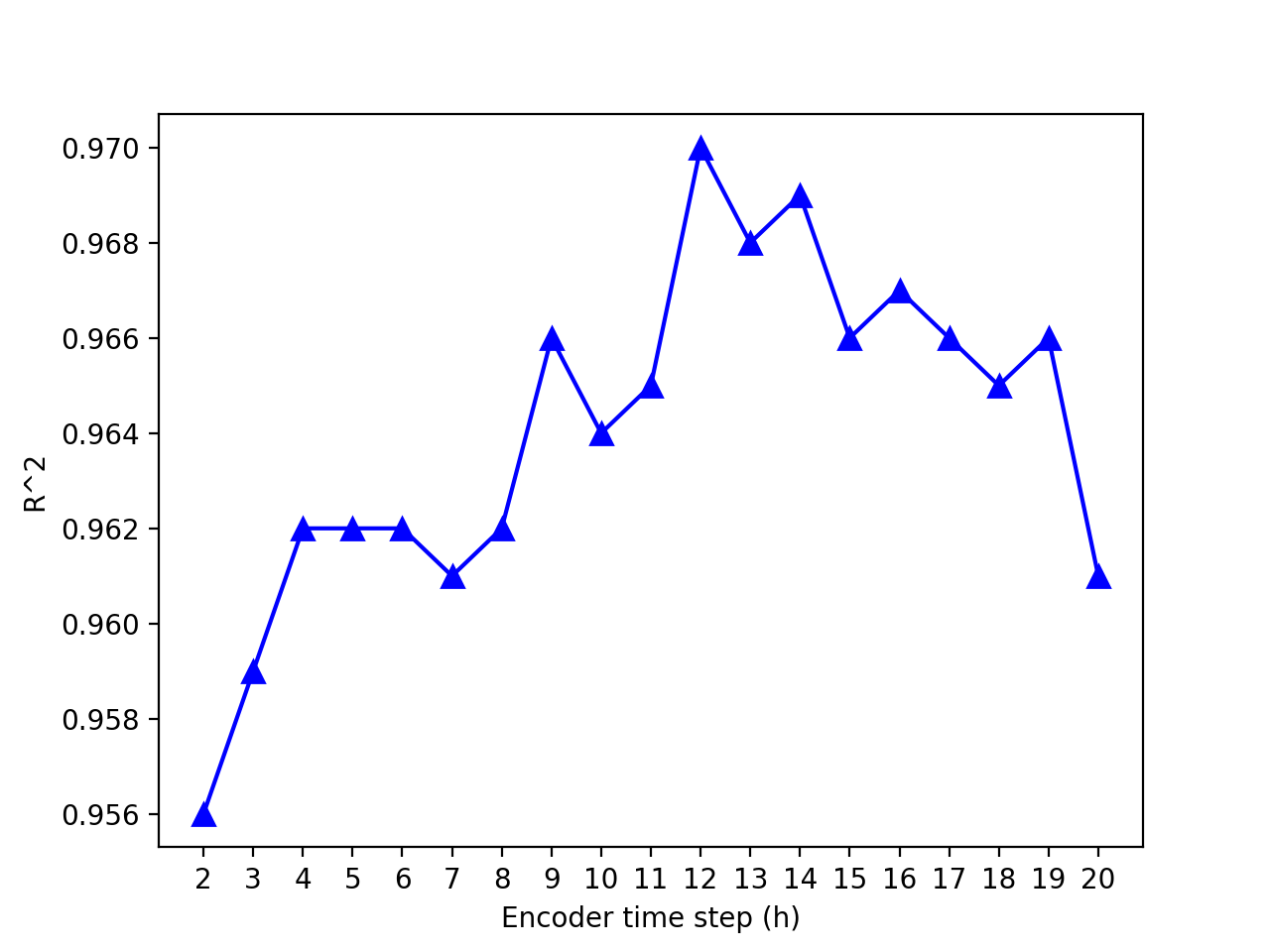}
\caption{R\textsuperscript{2}}
\end{center}
\label{etsr2}
\end{subfigure}
\caption{Errors of LSTM models with different encoder time steps}
\label{ETS}
\end{center}
\end{figure*}

The encoder time step is the number of hours of history data fed into the LSTM model. The encoder time step of the standard model in this paper is 12.
Fig.~\ref{ETS} shows the errors of models with different encoder time steps. Approximately, models with encoder time step in the range of [12,14] have the best forecast accuracy.

\begin{table}[t]
\caption{Comparison of Errors of Models}
\begin{center}
\begin{tabular}{ |c||c|c|c| }
 \hline
 Evaluation Criteria & SVR model & MLP model & LSTM model\\
 \hline
 RMSE & 136.022 & 99.359 & 82.007 \\
 \hline
 MAE & 63.939 & 35.248 & 27.752 \\
 \hline
 R\textsuperscript{2} & 0.917 & 0.956 & 0.970 \\
 \hline
\end{tabular}
\end{center}
\label{TabResult}
\end{table}

\begin{table*}[t]
\caption{Comparison of Errors of LSTM Models with Different Combinations of Input Data}
\begin{center}
    \begin{subtable}[t]{\textwidth}
    \begin{center}
        \begin{tabular}{|>{\centering\arraybackslash}m{3cm}||>{\centering\arraybackslash}m{1cm}|>{\centering\arraybackslash}m{1cm}|>{\centering\arraybackslash}m{2cm}|>{\centering\arraybackslash}m{3cm}|}
        \hline
         & None & Rainfall & Areal Rainfall & Rainfall + Areal Rainfall \\
        \hline
        With Stream-Flow & 148.864 & 82.007 & 94.008 & 85.772\\
        \hline
        Without Stream-Flow &  & 274.344 & 282.146 & 282.126\\
        \hline
       \end{tabular}
       \caption{RMSE}
       \label{input:RMSE}
    \end{center}
    \end{subtable}
    \begin{subtable}[t]{\textwidth}
    \begin{center}
        \begin{tabular}{|>{\centering\arraybackslash}m{3cm}||>{\centering\arraybackslash}m{1cm}|>{\centering\arraybackslash}m{1cm}|>{\centering\arraybackslash}m{2cm}|>{\centering\arraybackslash}m{3cm}|}
        \hline
         & None & Rainfall & Areal Rainfall & Rainfall + Areal Rainfall \\
        \hline
        With Stream-Flow & 43.188 & 27.752 & 29.751 & 30.138\\
        \hline
        Without Stream-Flow &  & 156.489 & 157.497 & 152.190\\
        \hline
       \end{tabular}
       \caption{MAE}
       \label{input:MAE}
    \end{center}
    \end{subtable}
    \begin{subtable}[t]{\textwidth}
    \begin{center}
        \begin{tabular}{|>{\centering\arraybackslash}m{3cm}||>{\centering\arraybackslash}m{1cm}|>{\centering\arraybackslash}m{1cm}|>{\centering\arraybackslash}m{2cm}|>{\centering\arraybackslash}m{3cm}|}
        \hline
         & None & Rainfall & Areal Rainfall & Rainfall + Areal Rainfall \\
        \hline
        With Stream-Flow & 0.900 & 0.970 & 0.960 & 0.967\\
        \hline
        Without Stream-Flow &  & 0.661 & 0.639 & 0.641\\
        \hline
       \end{tabular}
       \caption{R\textsuperscript{2}}
       \label{input:R2}
    \end{center}
    \end{subtable}
\label{tab:inputcombs}
\end{center}
\end{table*}

\section{Conclusion}

Forecast is always critical in saving human’s lives and properties from the flood disaster. This paper proposed a method of stream-flow forecast using LSTM network – a kind of deep learning neural network derived from RNN, equipped with a remember-forget system to avoid parameter blowing up or vanishing. To prove its advantage in time-series forecast with non-linear features, it is compared to the machine learning SVR model and deep learning MLP model in forecasting the stream-flow of Tunxi, China. Results of the experiment show that LSTM model provides more stable and more accurate prediction comparing to SVR and MLP models, proving its ability.

However, there is still room for improvement in the LSTM stream-flow forecasting model: the results show errors in peak volume forecast which cannot be ignored. The models may be improved in the following ways: First, Due to the limit of time and hardware capacity, the parameter choices of LSTM model are only based on simple tests and lack of thorough study. Moreover, most of default parameters of the model remain in their original value without adjustments. More study on the parameter adjustment may improve the model’s accuracy. Second, the data used in the experiment have a time span of more than 20 years. Due to the lack of technology and management in the past, the original data have a certain degree of disorder and deficiency, and various kinds of amendments are made to the data. Feeding data with higher quality may improve the model’s performance.

\bibliographystyle{./bibliography/IEEEtran}
\bibliography{./bibliography/PaperBib}

\begin{thebibliography}{10}
\providecommand{\url}[1]{#1}
\csname url@samestyle\endcsname
\providecommand{\newblock}{\relax}
\providecommand{\bibinfo}[2]{#2}
\providecommand{\BIBentrySTDinterwordspacing}{\spaceskip=0pt\relax}
\providecommand{\BIBentryALTinterwordstretchfactor}{4}
\providecommand{\BIBentryALTinterwordspacing}{\spaceskip=\fontdimen2\font plus
\BIBentryALTinterwordstretchfactor\fontdimen3\font minus
  \fontdimen4\font\relax}
\providecommand{\BIBforeignlanguage}[2]{{%
\expandafter\ifx\csname l@#1\endcsname\relax
\typeout{** WARNING: IEEEtran.bst: No hyphenation pattern has been}%
\typeout{** loaded for the language `#1'. Using the pattern for}%
\typeout{** the default language instead.}%
\else
\language=\csname l@#1\endcsname
\fi
#2}}
\providecommand{\BIBdecl}{\relax}
\BIBdecl

\bibitem{balica2013parametric}
S.~Balica, I.~Popescu, L.~Beevers, and N.~G. Wright, ``Parametric and
  physically based modelling techniques for flood risk and vulnerability
  assessment: a comparison,'' \emph{Environmental modelling \& software},
  vol.~41, pp. 84--92, 2013.

\bibitem{wallemacq2017disaster}
P.~Wallemacq, ``Natural disasters in 2017: Lower mortality, higher cost,''
  \emph{Brussels, Belgium: Centre for Research on the Epidemiology of
  Disasters}, 2018.

\bibitem{feng2018hydrologic}
J.~FENG and F.~PAN, ``A hydrologic forecast method based on lstm-bp,''
  \emph{Computer and Modernization}, no.~7, p.~19, 2018.

\bibitem{han2017bayesian}
S.~Han and P.~Coulibaly, ``Bayesian flood forecasting methods: A review,''
  \emph{Journal of Hydrology}, vol. 551, pp. 340--351, 2017.

\bibitem{damavandi2019accurate}
H.~G. Damavandi, R.~Shah, D.~Stampoulis, Y.~Wei, D.~Boscovic, and J.~Sabo,
  ``Accurate prediction of streamflow using long short-term memory network: A
  case study in the brazos river basin in texas,'' \emph{International Journal
  of Environmental Science and Development}, vol.~10, no.~10, 2019.

\bibitem{yaseen2015artificial}
Z.~M. Yaseen, A.~El-Shafie, O.~Jaafar, H.~A. Afan, and K.~N. Sayl, ``Artificial
  intelligence based models for stream-flow forecasting: 2000--2015,''
  \emph{Journal of Hydrology}, vol. 530, pp. 829--844, 2015.

\bibitem{haykin1994neural}
S.~Haykin, \emph{Neural networks: a comprehensive foundation}.\hskip 1em plus
  0.5em minus 0.4em\relax Prentice Hall PTR, 1994.

\bibitem{hsu2002self}
K.-l. Hsu, H.~V. Gupta, X.~Gao, S.~Sorooshian, and B.~Imam, ``Self-organizing
  linear output map (solo): An artificial neural network suitable for
  hydrologic modeling and analysis,'' \emph{Water Resources Research}, vol.~38,
  no.~12, pp. 38--1, 2002.

\bibitem{cigizoglu2005application}
H.~K. Cigizoglu, ``Application of generalized regression neural networks to
  intermittent flow forecasting and estimation,'' \emph{Journal of Hydrologic
  Engineering}, vol.~10, no.~4, pp. 336--341, 2005.

\bibitem{kagoda2010application}
P.~A. Kagoda, J.~Ndiritu, C.~Ntuli, and B.~Mwaka, ``Application of radial basis
  function neural networks to short-term streamflow forecasting,''
  \emph{Physics and Chemistry of the Earth, Parts A/B/C}, vol.~35, no. 13-14,
  pp. 571--581, 2010.

\bibitem{sivapragasam2005flow}
C.~Sivapragasam and S.-Y. Liong, ``Flow categorization model for improving
  forecasting,'' \emph{Hydrology Research}, vol.~36, no.~1, pp. 37--48, 2005.

\bibitem{asefa2006multi}
T.~Asefa, M.~Kemblowski, M.~McKee, and A.~Khalil, ``Multi-time scale stream
  flow predictions: The support vector machines approach,'' \emph{Journal of
  hydrology}, vol. 318, no. 1-4, pp. 7--16, 2006.

\bibitem{noori2011assessment}
R.~Noori, A.~Karbassi, A.~Moghaddamnia, D.~Han, M.~Zokaei-Ashtiani,
  A.~Farokhnia, and M.~G. Gousheh, ``Assessment of input variables
  determination on the svm model performance using pca, gamma test, and forward
  selection techniques for monthly stream flow prediction,'' \emph{Journal of
  Hydrology}, vol. 401, no. 3-4, pp. 177--189, 2011.

\bibitem{el2007neuro}
A.~El-Shafie, M.~R. Taha, and A.~Noureldin, ``A neuro-fuzzy model for inflow
  forecasting of the nile river at aswan high dam,'' \emph{Water resources
  management}, vol.~21, no.~3, pp. 533--556, 2007.

\bibitem{ozger2009comparison}
M.~{\"O}ZGER, ``Comparison of fuzzy inference systems for streamflow
  prediction,'' \emph{Hydrological Sciences Journal}, vol.~54, no.~2, pp.
  261--273, 2009.

\bibitem{sanikhani2012river}
H.~Sanikhani and O.~Kisi, ``River flow estimation and forecasting by using two
  different adaptive neuro-fuzzy approaches,'' \emph{Water resources
  management}, vol.~26, no.~6, pp. 1715--1729, 2012.

\bibitem{savic1999genetic}
D.~A. Savic, G.~A. Walters, and J.~W. Davidson, ``A genetic programming
  approach to rainfall-runoff modelling,'' \emph{Water Resources Management},
  vol.~13, no.~3, pp. 219--231, 1999.

\bibitem{makkeasorn2008short}
A.~Makkeasorn, N.-B. Chang, and X.~Zhou, ``Short-term streamflow forecasting
  with global climate change implications--a comparative study between genetic
  programming and neural network models,'' \emph{Journal of hydrology}, vol.
  352, no. 3-4, pp. 336--354, 2008.

\bibitem{guven2009linear}
A.~Guven, ``Linear genetic programming for time-series modelling of daily flow
  rate,'' \emph{Journal of earth system science}, vol. 118, no.~2, pp.
  137--146, 2009.

\bibitem{shiri2010short}
J.~Shiri and O.~Kisi, ``Short-term and long-term streamflow forecasting using a
  wavelet and neuro-fuzzy conjunction model,'' \emph{Journal of Hydrology},
  vol. 394, no. 3-4, pp. 486--493, 2010.

\bibitem{sahay2014predicting}
R.~R. Sahay and A.~Srivastava, ``Predicting monsoon floods in rivers embedding
  wavelet transform, genetic algorithm and neural network,'' \emph{Water
  resources management}, vol.~28, no.~2, pp. 301--317, 2014.

\bibitem{zhang2018developing}
J.~Zhang, Y.~Zhu, X.~Zhang, M.~Ye, and J.~Yang, ``Developing a long short-term
  memory (lstm) based model for predicting water table depth in agricultural
  areas,'' \emph{Journal of hydrology}, vol. 561, pp. 918--929, 2018.

\bibitem{pascanu2013construct}
R.~Pascanu, C.~Gulcehre, K.~Cho, and Y.~Bengio, ``How to construct deep
  recurrent neural networks,'' \emph{arXiv preprint arXiv:1312.6026}, 2013.

\bibitem{schmidhuber1997long}
J.~Schmidhuber and S.~Hochreiter, ``Long short-term memory,'' \emph{Neural
  Comput}, vol.~9, no.~8, pp. 1735--1780, 1997.

\bibitem{dong2015flood}
A.~Dong, L.~Zhi-Jia, W.~Yong-Tuo, Y.~Cheng, and D.~Yi-Heng, ``Flood forecasting
  model based on geographical information system,'' \emph{Proceedings of the
  International Association of Hydrological Sciences}, vol. 368, pp. 192--196,
  2015.

\bibitem{basak2007support}
D.~Basak, S.~Pal, and D.~C. Patranabis, ``Support vector regression,''
  \emph{Neural Information Processing-Letters and Reviews}, vol.~11, no.~10,
  pp. 203--224, 2007.

\bibitem{bourlard1988auto}
H.~Bourlard and Y.~Kamp, ``Auto-association by multilayer perceptrons and
  singular value decomposition,'' \emph{Biological cybernetics}, vol.~59, no.
  4-5, pp. 291--294, 1988.

\end{thebibliography}

\end{document}